%% file: emnlp2022.tex
\pdfoutput=1

\documentclass[11pt]{article}

\usepackage[]{EMNLP2022}

\usepackage[T1]{fontenc}

\usepackage[utf8]{inputenc}

\usepackage{microtype}

\usepackage{inconsolata}

\usepackage{times}
\usepackage{latexsym}
\usepackage{amssymb}
\usepackage{amsmath}
\usepackage{algorithm}
\usepackage{algorithmicx}
\usepackage{algpseudocode}
\usepackage{tabularx}
\usepackage{booktabs}
\usepackage{multicol}
\usepackage{multirow}
\usepackage{xcolor}
\usepackage{xspace}

\usepackage[T1]{fontenc}

\usepackage[utf8]{inputenc}

\usepackage{microtype}

\usepackage{graphicx}
\usepackage{subfig}

\newcommand{\logicnlg}{\textsc{LogicNLG}\xspace}
\newcommand{\plog}{\textsc{PLoG}\xspace}
\newcommand{\logictext}{\textsc{Logic2text}\xspace}

\newcommand{\tapex}{\textsc{TaPEx}\xspace}
\newcommand{\contlog}{\textsc{ContLog}\xspace}
\newcommand{\tapas}{\textsc{TaPas}\xspace}

\title{\textsc{PLoG}:  Table-to-Logic Pretraining for Logical Table-to-Text Generation}

\author{Ao Liu$^1$\thanks{{} {} Work done during internship at MSRA.},  Haoyu Dong$^2$\thanks{{} {} Corresponding author.}, Naoaki Okazaki$^1$, Shi Han$^2$, Dongmei Zhang$^2$ \\
      $^1$ Tokyo Institute of Technology,  
      $^2$ Microsoft Research\\
      \texttt{liu.ao@nlp.c.titech.ac.jp}, \texttt{okazaki@c.titech.ac.jp} \\
      \texttt{\{hadong,shihan,dongmeiz\}@microsoft.com} 
      }
\begin{document}
\maketitle

\input{Abstract}

\input{Introduction}

\input{RelatedWork}

\input{Method}

\input{Experiment}

\input{Conclusion}

\section*{Limitations}
The first limitation of our approach is that it is initialized from pretrained language models such as T5 to inherit the language generation knowledge learned from large-scale text corpora. This requires the input of \plog to be a text sequence, which may limit the structural encoding of table inputs and logical form outputs. Although it is possible for us to design and pretrain a new model from scratch, the computational cost will be too large. The second limitation is also caused by this. Because we adopt pretrained language models to perform table-to-logic and table-to-text generation, we have to serialize the input (sub-) tables to fit them in the language model encoder. Therefore, the maximum sequence length of the encoder model limited the size of the input table. To address this, we only input relevant columns or highlighted cells instead of the full table to reduce the input sequence length. However, some potentially useful contextual information in the full table is omitted and may limit the model performance. The third limitation lies in the logical form schema we adopt, which is restricted to the domain of current logical table-to-text datasets. When applying our method to new downstream datasets with unseen logic types, e.g., median, proportion, the current schema should be extended to support the new logic. However, the schema is easy to extend by defining new logical operations as executable functions on tables. 

\section*{Ethics Statement}
This work presents \plog, a pretrained language model for the research community to study logical table-to-text generation. In addition, we also propose a new dataset \contlog for the research of controlled logical table-to-text generation. Our dataset contains Wikipedia tables, annotations (target sentences, meta information such as logic types) and highlighted table cell information. We reuse the tables and annotations of \logictext. \logictext is a public dataset under MIT
license. And to obtain the highlighted cell information, we use an automatic method without human annotation. We also use \logicnlg, another public dataset for experiments, which is also under MIT license. All datasets are in English.
In human evaluation, we hire human annotators to evaluate the performance of our models. We recruit 3 graduate students in electrical engineering, computer science, and English majors (1
female and 2 males). Each student is paid \$7.8 per hour (above the average local payment of similar jobs).

\section*{Acknowledgements}
This work was supported by JST, the establishment of university fellowships towards the creation of science technology innovation, Grant Number JPMJFS2112.

\bibliography{custom}
\bibliographystyle{acl_natbib}
\clearpage

\input{Appendix}

\end{document}

%% file: Abstract.tex
\begin{abstract}
Logical table-to-text generation is a task that involves generating logically faithful sentences from tables, which requires models to derive logical-level facts from table records via logical inference. It raises a new challenge on the logical-level content planning of table-to-text models. However, directly learning the logical inference knowledge from table-text pairs is very difficult for neural models because of the ambiguity of natural language and the scarcity of parallel data. Hence even large-scale pre-trained language models present low logical fidelity on logical table-to-text. In this work, we propose a Pretrained Logical Form Generator (\plog) framework to improve generation fidelity. Specifically, \plog is first pretrained on a table-to-logical-form generation (\textit{table-to-logic}) task, then finetuned on downstream table-to-text tasks. The logical forms are formally defined with unambiguous semantics. Hence we can collect a large amount of accurate logical forms from tables without human annotation. In addition, \plog can learn logical inference from table-logic pairs much more reliably than from table-text pairs. To evaluate our model, we further collect a controlled logical table-to-text dataset \contlog based on an existing dataset. On two benchmarks, \logicnlg and \contlog, \plog outperforms strong baselines by a large margin on logical fidelity, demonstrating the effectiveness of table-to-logic pretraining.

\end{abstract}

%% file: Introduction.tex
\begin{figure*}[t]
    \centering
    \includegraphics[width=\linewidth]{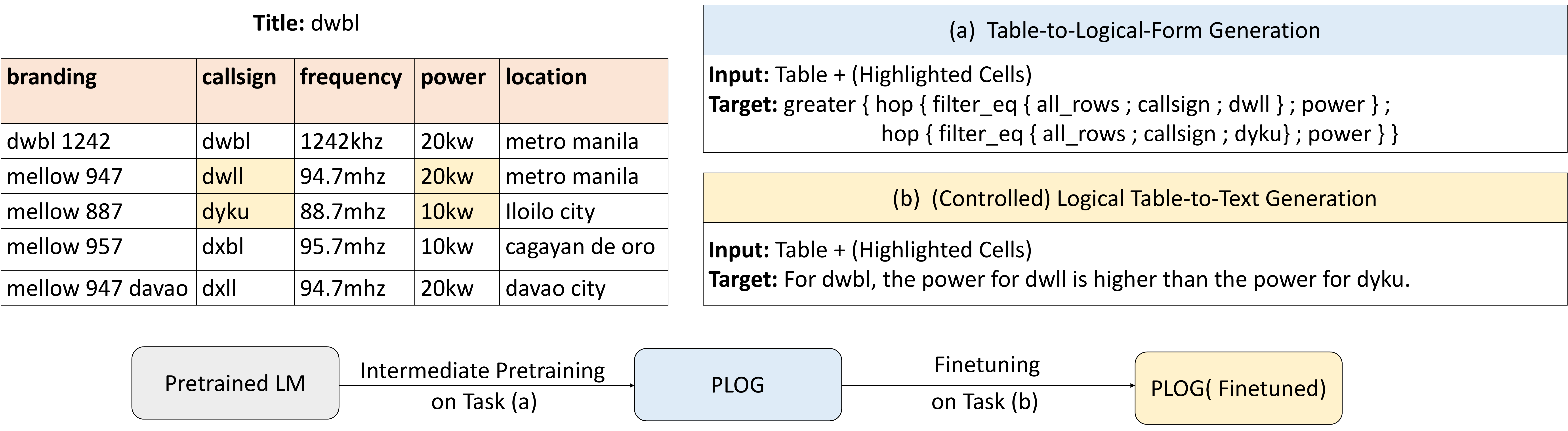}
    \caption{Examples of the tasks and the training procedure of our proposed \plog model. Task (a) is the table-to-logic pretraining task we propose; task (b) is the downstream logical table-to-text task we target. The yellow-colored table cells are annotated as control features for the \contlog task, while for \logicnlg, such highlighted cells are not available. We collect different table-to-logic datasets for \contlog and \logicnlg separately and perform intermediate pretraining for pretrained language models on the collected data, then finetune the model on the downstream tasks.}
    \label{fig:main}
\end{figure*}

\section{Introduction}
\label{sec:intro}
Table-to-text generation is a sub-task of data-to-text generation, aiming to generate natural language descriptions from structured tables. There are two main steps to performing table-to-text generation: content planning (selecting table contents and determining the plan to describe them) and surface realization (verbalizing the plan into fluent natural language). Traditional table-to-text systems adopt a pipeline architecture to complete the two procedures with separate modules~\cite{kukich1983design, mckeown1985discourse}. Recent work has shown the advantage of using a neural encoder-decoder model to directly generate sentences from the tables, which presents the strong capability to produce fluent and natural text~\cite{wiseman2017challenges, nie2018operation, puduppully2019entity}. Researchers have also attempted to finetune pretrained language models such as BART~\cite{bart} and T5~\cite{2020t5} on downstream table-to-text tasks and achieved remarkable success on a broad range of benchmarks~\cite{xie2022unifiedskg, kale2020text}. 

Previous studies have mainly focused on surface-level realization, i.e., simply restating surface-level facts in natural language~\cite{wiseman2017challenges, liu2018table, puduppully2019data, puduppully2019entity}. Recently, logical table-to-text generation~\cite{chen2020logical}, i.e., generating textual descriptions that require logical reasoning over surface-level facts in the table, has attracted increasing attention. Logical table-to-text generation poses a new challenge of \textit{logical-level} content planning, requiring models to perform logical inference to derive facts from surface-level table records. End-to-end neural models often suffer from low logical fidelity on this task, i.e., the generated sentences are not logically entailed by the tables despite their reasonable fluency~\cite{chen2020logical, chen2021confounded}. We attribute this to the fact that the ambiguity of natural language target sentences hinders neural models from learning accurate logical inference from table-text pairs. In addition, the amount of such table-text pairs is limited because of the labor-intensive human annotation for logic-focused descriptions, which also limits the performance of neural models.

To achieve high fidelity of logical-level generation, \citet{chen2020logic2text} have attempted to annotate logical forms to guide the text generation and proposed a \textsc{Logic2text} dataset. With logical forms as mediators conveying accurate logical-level facts, models can focus on surface realization from associated logical forms and achieve high fidelity. However, annotating accurate logical forms for textual descriptions requires intensive human efforts. Moreover, generating from a self-contained logical form is actually a different task from table-to-text generation. Prior studies on this dataset~\cite{liu2021improving, shu2021logic, xie2022unifiedskg} mostly focus on converting the logical forms into texts rather than tables into texts.

In this study, we propose a \textbf{\textsc{P}}re-trained \textbf{\textsc{Lo}}gical Form \textbf{\textsc{G}}enerator (\textbf{\textsc{PLoG}}) model to achieve more faithful logical table-to-text.  Specifically, \textsc{PLoG} is first pretrained on a large-scale synthetic corpus of table-to-logical-form generation (\textit{table-to-logic}) to learn how to generate accurate logical forms from tables, then finetuned on downstream table-to-text tasks to transfer the logical inference knowledge learned from pretraining to text generation. Our insights are three-fold. (i) Unlike natural language sentences, logical forms are formally defined with unambiguous semantics; hence it is much easier and more reliable for models to acquire logical inference knowledge via learning from logical form generation. (ii) It is viable to collect large-scale logical form corpora via rule-based sampling over tables without the efforts of human annotators. (iii) Via pretraining on large amounts of table-to-logic data, the proposed model can better understand the table and organize the logical-level content planning, leading to faithful table-to-text generation. Here, we treat logical forms as intermediate meaning representations of logical-level texts, while we do not need them when performing the downstream task. To collect the pretraining data, we propose an execution-guided sampling approach to sample accurate logical forms from tables automatically.

We formulate the pretraining task in the same sequence-to-sequence (seq2seq) generation to achieve smooth transfer learning to the downstream table-to-text task. We adopt several strong pretrained language models, BART and T5, as the backbone models. Because the previous benchmark for logical table-to-text, \logicnlg, lacks control features, leading to uncontrollable content selection and poor logical fidelity, we collect a a new \textbf{\textsc{Cont}}rolled \textbf{\textsc{Log}}ical Natural Language Generation (\contlog) dataset as a complementary testbed towards controlled logical table-to-text generation. Specifically, we re-organize the \logictext dataset by detecting highlighted cells based on their annotated logical forms. Figure~\ref{fig:main} presents examples of the table-to-logic pretraining task and the (controlled) logical table-to-text task.

On the two benchmarks, \logicnlg and \contlog, \plog outperforms the strong baselines such as T5 by a large margin on the logical fidelity, demonstrating the effectiveness of table-to-logic pretraining. Human evaluation and analysis experiments further demonstrate that our approach can considerably promote the fidelity of logical table-to-text generation.\footnote{Code and data will be available at \url{https://github.com/microsoft/PLOG} after passing an internal compliance review.}

%% file: RelatedWork.tex
\section{Related Work}
\paragraph{Table-to-Text Generation} Early table-to-text generation tasks are limited to surface-level generation with little focus on logical inference~\cite{lebret2016neural}. \logicnlg is the first dataset to focus on logical table-to-text generation, with Wikipedia tables and human-annotated logical descriptions. \citet{chen2021confounded} proposed a de-confounded variational encoder-decoder model to encourage the model to generate non-surface-level predictions; however, the logical reasoning process is not explicitly considered, leading to low fidelity scores on human evaluation. \citet{chen2020logic2text} proposed to annotate logical forms to guide the generation and released a \logictext dataset. In this work, we focus on direct logical table-to-text generation without any explicit logical forms. 
Another related line of datasets are ToTTo~\cite{parikh2020totto} and HiTab~\cite{cheng2021hitab}, which incorporate highlighted cells to promote controllable generation. The \contlog dataset we propose is similar to the task settings of these datasets but differs in that we focus on logical-level generation. At the same time, only a small portion of examples in ToTTo and HiTab involve logical reasoning. ROTOWIRE~\cite{wiseman2017challenges} and NumericNLG~\cite{suadaa2021numericnlg} also involve numerical reasoning over table records, while they focus on document-level table summarization instead of sentence generation.

\paragraph{Table Pretraining} Table pretraining~\cite{tapas2020, liu2021tapex,dong2022table, iida2021tabbie} has been popular for table understanding tasks such as Table Question Answering (TableQA)~\cite{wikisql, pasupat2015wikitable} and Table Fact Verification (TableFV)~\cite{chen2019tabfact}. With large-scale pretraining corpora, the table pretraining models can learn a better joint understanding of tabular and textual data through well-defined pretraining objectives. Most table pretraining works are based on table-text corpora, while \tapex~\cite{liu2021tapex} learns from synthetic SQL programs, which is the closest to our work. Specifically, \tapex is first pretrained on a table-based SQL execution task, where the input is a table and a SQL program, and the output is the answer to the SQL query. Then, the pretrained model can be finetuned on TableQA and TableFV tasks where the input is a table associated with a textual query/statement, and the output is the answer. However, our work differs from \tapex in that 
we focus on table-to-text generation, where the input is a structured table and the output is a textual statement of the table contents. Our task requires deriving a complete logical-level fact from the table without the guidance of any query. In addition, our pretraining task also requires generating a self-contained logical form from the table, while \tapex aims to learn the neural execution of an existing SQL program. Similarly, FLAP~\cite{flap2021} proposes to enhance the numerical reasoning ability of table-to-text models with an artificial pretraining task. This task is a synthetic QA task similar to \tapex pretraining. 

Another line of related works adopts pretraining techniques to solve the text-to-SQL parsing~\cite{yu2021grappa, shi2021gap} task, also involving collecting synthetic SQL data and pretraining models on SQL generation tasks. However, text-to-SQL still requires an explicit NL query as the input, which is different from our task. Although table pretraining is popular in table understanding tasks, it has not been well-explored in table-to-text. Previous works on table-to-text tend to directly utilize pretrained language models by flattening structured tables into sequences~\cite{gong2020tablegpt, kale2020text, xie2022unifiedskg}. A recent work~\cite{tabt5} incorporates structural positional embeddings of tables into T5~\cite{2020t5} and performs intermediate pretraining in a similar way to \tapas~\cite{tapas2020}. Similarly, \plog can also be seen as intermediate pretraining of language models for table-to-text generation.

%% file: Method.tex
\section{Downstream Tasks}
In this work, we focus on logical table-to-text. The previous benchmark \logicnlg aims at generating sentences from a full table without control features, which causes uncontrollable content selection and hinders faithful generation~\cite{chen2020logic2text}. Therefore, we propose a new controlled logical table-to-text dataset \contlog as a complementary testbed to \logicnlg. Inspired by previous studies on controlled table-to-text~\cite{parikh2020totto, cheng2021hitab}, we incorporate highlighted cells as additional supervision signals in \contlog (Figure~\ref{fig:main}) to narrow down the scope of content selection, such that models can focus more on planning and generation. 

\subsection{\contlog Dataset Construction}
 We reuse the \logictext dataset to build \contlog. In \logictext, there is an annotated logical form for each target sentence. The logical form can convey the accurate logical semantics of the sentence. Hence, we execute the logical forms on the context tables and extract the table cells relevant to the execution process. These cells are also the ones most relevant to the target sentence. Although built upon \logictext, \contlog does not contain logical forms because we focus on the direct table-to-text generation. Figure~\ref{fig:main} shows an example of \contlog.

\subsection{Task Formulation}
\label{sec:task}
The input of \logicnlg is a table $T$ with an NL title $W$.  $T = \{T_{ij} | 1 \leq i \leq R_T, 1 \leq j\leq R_C\}$, where $R_T$ and $R_C$ are the numbers of rows and columns, respectively, and $T_{ij}$ is the table cell value at row $i$ and column $j$. Each column also has a column header $Col_j$. The output is a sentence $y$. The task objective is to find a model $P(y|T)$ to generate a sentence $\hat{y}$ that is both fluent and logically entailed by the table. In \contlog, an additional set of highlighted cells $H=\{(i,j)\}$ are included as a part of the input, where $i$ and $j$ denote the row index and column index of a highlighted cell. The objective thus becomes $P(y|T; H)$.

\section{Table-to-Logic Pretraining}
Logical table-to-text is difficult mainly because of the ambiguity of natural language sentences. For example, a sentence \texttt{Alice was the first player that achieved champion in 2010} has two possible meanings: (1) \textit{Alice got the first champion of 2010}; (2) \textit{Alice became the first champion in history, and this achievement happened in 2010.} This prevents end-to-end neural models from inferring unambiguous logical facts from the table, especially when the parallel data are scarce.

To achieve faithful logical table-to-text generation, we propose a table-to-logic pretraining task that involves generating a logical form from an input table. In this task, the model needs to mine logical-level facts from tables and organize the facts into formally defined meaning representations, i.e., logical forms. Each logical form can be regarded as an abstract content plan of a logical-level description. Therefore, we expect a model to learn logical-level content planning from the pretraining task. We then finetune the model on the downstream table-to-text tasks to generalize the content planning to natural language generation. We formulate our pretraining and downstream tasks as the same seq2seq generation paradigm to realize successful transfer learning.

\subsection{Pretraining Task Formulation}
The input of the pretraining task is the same (sub-) table as we introduced in Section~\ref{sec:task}, while the target is a logical form instead of a sentence. We follow the same schema in \logictext to define the logical forms used in our task. Each logical form $z$ is the composition of several logical functions. Each function $f_i(arg_1, ...)$ accepts several arguments relevant to the table $T$. $z$ can be parsed into a tree and executed from bottom to up by a logical form executor. In this process, the execution result of $f_i$ may be fed to its parent function as an argument. The root function always outputs a Boolean value (\textit{true} or \textit{false}) which indicates the factual correctness of $z$. We select this schema because of its several merits. (1) It is originally designed to represent logical-level textual statements in \logictext, and its definition is close to our downstream tasks. A similar schema has also been used for TableFV tasks~\cite{ou-liu-2022-learning, chen2019tabfact}. (2) It covers seven of the most commonly used logic types: \texttt{count, unique, comparative, superlative, ordinal, aggregation} and \texttt{majority}. (3) The logical forms can be executed on the tables to evaluate their exact correctness, allowing accurate evaluation of the pretraining task.
A detailed description of the logic schema is provided in Appendix~\ref{sec:schema}.

\begin{figure*}[t]
    \centering
    \includegraphics[width=0.9\linewidth]{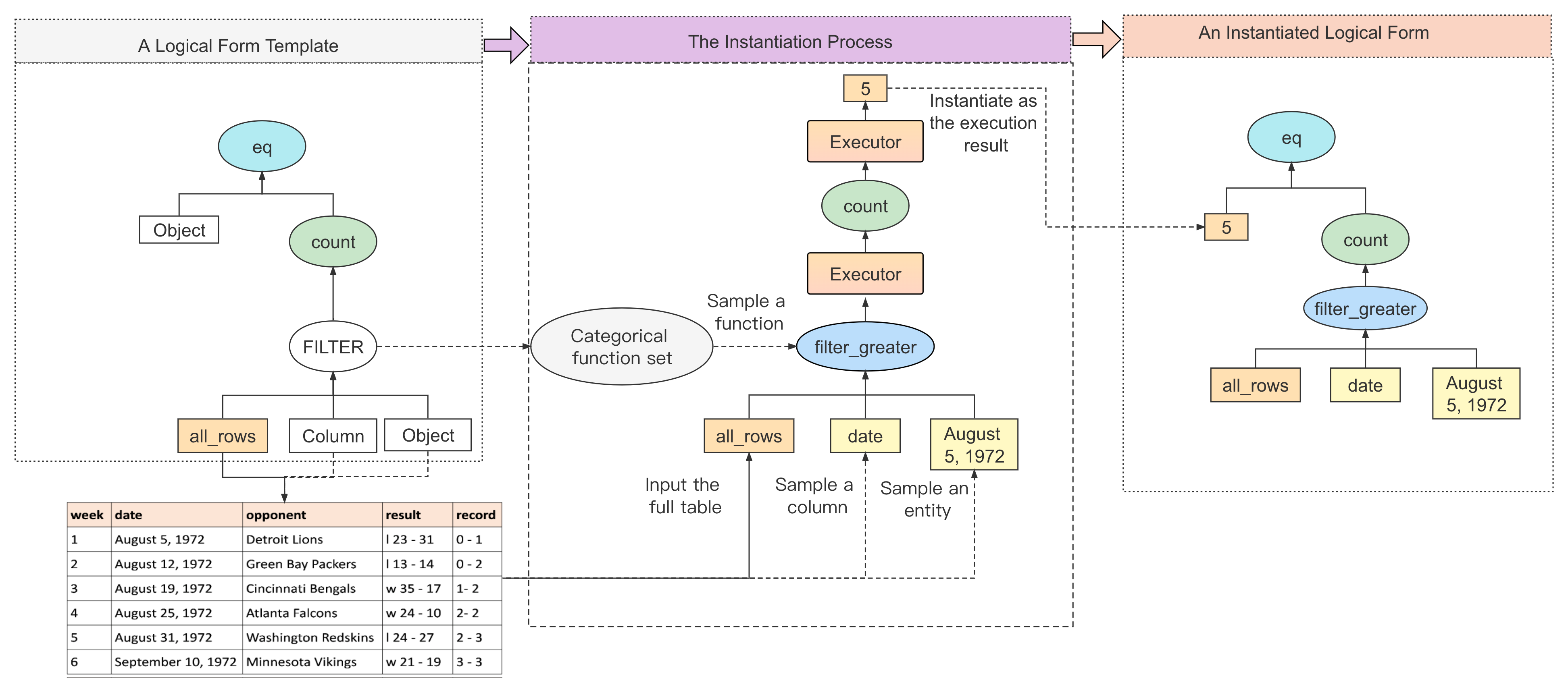}
    \caption{An example of instantiating a logical form template. The colored nodes in the template indicate nodes that do not need instantiation, while the white-background nodes are typed placeholders in the template. The dotted arrows indicate instantiation. We employ instantiation of these white nodes with a bottom-up execution-guided sampling approach. Finally, a logical form instance is obtained. \texttt{Column} and \texttt{Object} indicate a column header and an object (entity/number), respectively. \texttt{FILTER} indicates the category of row-filtering functions. \texttt{all\_rows} is a special entity to represent the entire table. }
    \label{fig:inst}
\end{figure*}

\subsection{Evaluation Metric of Table-to-Logic} We adopt the execution accuracy of generated logical forms as the evaluation metric for our pretraining task, similar to the setting in text-to-SQL tasks~\cite{wikisql}. Specifically, a logical form is counted as correct if it can be successfully executed on the input table and returns a Boolean value \textit{True} that indicates the table entails it. 

\subsection{Pretraining Data Collection}
\label{sec:pt-data}
To perform table-to-logic pretraining, we must collect enough paired data of tables and associated logical forms. The formal definition of logical forms allows us to automatically collect a large amount of logical forms from tables via rule-based sampling. Here, we propose instantiating existing logical form templates to sample logical forms similarly to how prior studies collect SQL programs~\cite{zhong2020grounded, liu2021tapex}. Specifically, we extract abstract templates from the logic schema we use. Then we adopt an execution-guided sampling method to instantiate the templates based on the context tables. Our approach has two merits: (1) By utilizing the pre-defined templates, we can control the distribution and diversity of collected logical forms. (2) With the execution-guided sampling, the correctness of the collected logical forms is guaranteed.

\paragraph{Templatization} 

We first extract the templates based on our logic schema. We define them as trees with typed placeholder nodes that need to be instantiated into specific functions or entities.
The placeholders include two entity types: \texttt{Column} represents a column header and \texttt{Object} means either a textual entity or a numerical value. In addition, we categorize some similar functions into smaller groups to obtain several function placeholders, which can reduce the number of templates and simplify the instantiation work. For example, \texttt{FILTER} represents a set of row-filtering functions. Table~\ref{tab:funcset} shows the complete list of these function placeholders. The other functions that cannot be categorized need not instantiation.
Finally, we obtain 35 templates, an average of 5 for each logic type. More examples of the templates are provided in Appendix~\ref{sec:collect}.

\paragraph{Instantiation} We propose an execution-guided bottom-up sampling strategy to instantiate the template trees. An example of template instantiation is depicted in Figure~\ref{fig:inst}. We design rules to instantiate different placeholder nodes via sampling. For example, we uniformly sample a column from the table to instantiate a \texttt{Column} placeholder (e.g. \texttt{date} in Figure~\ref{fig:inst}). For a function placeholder such as \texttt{FILTER}, we sample a specific function from the corresponding category it represents (e.g. \texttt{filter\_greater} in Figure~\ref{fig:inst}). For each instantiated function node, we execute it, obtain the execution result and feed the result to the parent function as an argument. Hence, the arguments of higher-level functions are guaranteed to be valid. The process lasts from bottom to up until finishing executing the root function node. We provide the detailed sampling rules in Appendix~\ref{sec:collect}.
 For each table, we conduct multiple trials of sampling. At each trial, we randomly select a template based on its distribution in \logictext, and perform the instantiation. Because of the randomness in selecting functions and entities, we can obtain different results from multiple trials. A trial may sometimes fail because of execution errors, but each successful trial will result in a correct logical form. We can perform the sampling as many trials as we want to obtain a large-scale table-to-logic corpus.

\paragraph{Table Source and Data Collection}
We collect pretraining data separately for the two datasets, \logicnlg and \contlog. For each dataset, we use the tables in its training data as the source tables to avoid potential data leakage. In addition, we remove the sampled logical forms that have appeared in \logictext since they are semantically equal to some of the target sentences in \contlog. To evaluate the performance of table-to-logic models and enable the selection of pretrained models, we also split the collected logical forms into train/val/split sets. The statistics of the pretraining data and their corresponding downstream datasets are shown in Table~\ref{tab:pt-stat}. Although we can sample more logical forms with more trials, we find the current pretraining data enough to obtain ideal experimental results.

\begin{table}[t]
    \centering
    \resizebox{\linewidth}{!}{
    \begin{tabular}{lccc}
    \toprule
    Dataset & \#tables  & \#examples (train/val/test) \\
    \midrule
     \logicnlg &  7,392 & 28,450/4,260/4,305 \\
     \contlog   & 5,554 &  8,566/1,095/1,092 \\ \midrule
    \logicnlg (pretrain)   & 5,682  &
    426.6k/3,000/2,997 \\
    \contlog (pretrain)   & 4,554 & 800k/1,500/1,500\\
    \bottomrule
    \end{tabular}}
    \caption{Statistics of the downstream tasks and their corresponding table-to-logic pretraining data.}
    \label{tab:pt-stat}
\end{table}

\section{The \plog Model}
\label{sec:model}
In this section, we introduce our proposed model \plog and how we conduct the seq2seq generation for the pretraining and downstream tasks.
\paragraph{Backbone Model}
We utilize the same backbone model to address both tasks to achieve the knowledge transfer from the table-to-logic pretraining task to the table-to-text downstream task. Theoretically, any text generation model applies to our task, such as GPT-2~\cite{gpt2}, BART, and T5. We test different backbone models, including BART-large, T5-base, and T5-large.

\paragraph{Model Input} Similarly to prior work on table-to-text generation~\cite{kale2020text, parikh2020totto}, we employ a template-based method to serialize the input table. For the \logicnlg task, we follow \cite{chen2020logical} to encode the relevant table columns by concatenating the table cells in row-wise order. For \contlog, we only concatenate the highlighted table cells as the input, as suggested by prior works on controlled table-to-text generation~\cite{parikh2020totto}.
This is to avoid the over-length issue with pretrained models and the negative impacts caused by irrelevant table information.

\paragraph{Numerical Pre-Computation} Numerical reasoning is difficult for neural language models, especially aggregation operations (e.g., the average of numerical values) and numerical ranking (e.g., the nth-maximum values of a column). Therefore, we conduct a pre-processing step by pre-computing some potentially useful numerical values. Similar approaches have also been proposed to improve the fidelity in table summarization~\cite{suadaa2021numericnlg} and text-to-SQL tasks~\cite{zhao2022bridging}. First, we evaluate each numerical cell's rank in its column (or the scope of highlighted cells) and append this rank to the linearized cell representation. Hence, each table cell $T_{ij}$ can be serialized into a sequence $c_{ij}=$ \texttt{<cell> $T_{ij}$ <col\_header> $Col_j$ </col\_header> <row\_idx> $i$ </row\_idx> <max\_rank> $r^+_{ij}$ </max\_rank> <min\_rank> $r^-_{ij}$ </min\_rank> </cell>}, where $r^+_{ij}$ indicates the rank of $T_{ij}$ in column $j$ in the decreasing order and $r^-_{ij}$ is the rank in the increasing order. The special tokens with angle brackets are used to indicate the structure of the input. In addition, we compute the average and sum of each numerical column in the input (sub-) table, and append two aggregation cell strings $c_{sum}$ and $c_{avg}$ to the flattened table sequence. $c_j^{sum}/c_j^{avg}$ = \texttt{<sum\_cell>/<avg\_cell> sum\_value/avg\_value <col\_header> $Col_j$ </col\_header> </sum\_cell>/</avg\_cell>}.

Finally, the input (sub-) table is serialized as $S$ = \texttt{<table> <caption> $W$ </caption> $c_1^{sum}$ $c_1^{avg}$ ... $c_{11}$ $c_{12}$ ... </table>}.

\paragraph{Model Output} We linearize each logical form $z$ into a string via a pre-order traversal of the logic tree following \cite{chen2020logic2text}. Special punctuations such as semicolons and braces are used to indicate the structural relationships between functions. For example, the logical form instance in Figure~\ref{fig:inst} can be linearized into \texttt{eq \{ 5 ; count \{ filter\_greater \{ all\_rows ; date ; August 5, 1972 \} \} \}}. As for the downstream task, the output becomes a sentence. After pretraining a \plog model, we directly finetune it on the downstream table-to-text tasks by changing the target from logical forms to sentences.

%% file: Experiment.tex
\section{Experiments}

\begin{table*}[!t]
\centering
\resizebox{\linewidth}{!}{
\begin{tabular}{lccccccc}
\toprule
\multirow{2}{*}{\textbf{Model}}   & \multicolumn{3}{c}{\textbf{Surface-level Evaluation}} & \multicolumn{4}{c}{\textbf{Logical Fidelity}} \\
 \cmidrule(lr){2-4}  \cmidrule(lr){5-8} 
 & BLEU-1 & BLEU-2 & BLEU-3 & SP-Acc & NLI-Acc & \tapex-Acc & \tapas-Acc \\

\midrule
 
GPT-TabGen (sm)     & 48.8  & 27.1   & 12.6        & 42.1 & 68.7  & 46.0 &45.5 \\

GPT-Coarse-to-Fine (sm)   & 46.6 & 26.8 & 13.3 & 42.7  & 72.2  &44.6 & 45.6\\

DCVED + GPT-TabGen  & 49.5 & 28.6 & 15.3 &  43.9 & 76.9 & -- &  --\\
\midrule

T5-base  & \textbf{52.6} &\textbf{32.6} &	\textbf{19.3} & 	48.2 & 80.4 & 52.4 & 56.2\\
\plog (T5-base)  & 51.7 &	32.3 &	18.9 & \textbf{48.9} & \textbf{85.5} &\textbf{61.7} & \textbf{62.3} \\ 
\midrule
T5-large &  53.4 & \textbf{34.1} & \textbf{20.4} & 48.4 &85.9 & 65.5 & 66.2 \\
\plog (T5-large) & \textbf{53.7} & \textbf{34.1} & \textbf{20.4} & \textbf{54.1} & \textbf{89.0} & \textbf{75.9} &\textbf{76.0} \\
\midrule
BART-large  &54.5 & 34.6 & 20.6 & 49.6 & 85.4 & 63.3 & 67.1\\
\plog (BART-large)  & \textbf{54.9} & \textbf{35.0} & \textbf{21.0} & \textbf{50.5} & \textbf{88.9} & \textbf{73.7} & \textbf{74.4} \\

\bottomrule
\end{tabular}}
\caption{The experimental results of different models on the test split of \logicnlg. For the previous models, we compute the \tapex-Acc and \tapas-Acc of the only two that have a released official output. We compare each pair of base and \plog models and mark the better scores as bold.}
\label{tab:lnlg-res}
\end{table*}

\begin{table*}[t]
    \centering
    \small
    \resizebox{\linewidth}{!}{
    \begin{tabular}{lccccccc}
    \toprule
   \multirow{2}{*}{\textbf{Model}}  & \multicolumn{5}{c}{\textbf{Surface-level Evaluation}} & \multicolumn{2}{c}{\textbf{Logical Fidelity}} \\
 \cmidrule(lr){2-6}  \cmidrule(lr){7-8} 
        &  BLEU-4 & ROUGE-1 & ROUGE-2 &ROUGE-4 & ROUGE-L & \tapex-Acc & \tapas-Acc \\
        \midrule
    
   T5-base  & 29.7	& 60.2	& 36.4	& 16.4 &	50.2 &	67.4 & 64.8 \\
    \plog (T5-base) & \textbf{30.4} &	\textbf{61.4} &	\textbf{37.3} &	\textbf{16.8} &	\textbf{51.4} &	\textbf{78.3} & \textbf{74.0} \\
    \midrule

  T5-large  & 31.2 & 62.1 & 37.9 & \textbf{17.6} & 51.4 &73.8 & 71.3  \\
  \plog (T5-large) & \textbf{31.7} & \textbf{62.3} & \textbf{38.3} & \textbf{17.6} & \textbf{52.0} & \textbf{81.9} & \textbf{76.8}\\
  \midrule
  
   BART-large & 29.3 &59.6 & 36.0 & 16.3 & 48.9 & 70.3 & 64.8 \\
    \plog (BART-large) & \textbf{32.1} & \textbf{63.2} & \textbf{39.2} & \textbf{18.1} & \textbf{53.0} & \textbf{85.9} & \textbf{82.0}\\
    \bottomrule
    \end{tabular} }
    \caption{The experimental results of different models on the test split of \contlog. We compare each pair of base and \plog models and mark the better scores as bold.}
    \vspace{-2ex}
    \label{tab:clog-res}
\end{table*}

\begin{table}[t]
    \centering
    \small
    \begin{tabular}{lcccc}
    \toprule
    \multirow{2}{*}{Model}     & \multicolumn{2}{c}{\logicnlg} & \multicolumn{2}{c}{\contlog} \\
    \cmidrule(lr){2-3} \cmidrule(lr){4-5}
    & AVG & ACC & AVG & ACC\\
    \midrule

   T5-base     & \textbf{1.87} & \textbf{40.5\%} &  2.15  & 58.0\% \\
   \plog (T5-base) & 1.84 & 40.0\% & \textbf{2.42} & \textbf{71.5\%} \\
   \midrule
   T5-large     & 2.21 & 55.0\% &  2.42  & 70.5\% \\
   \plog (T5-large) & \textbf{2.41} & \textbf{66.0\%} & \textbf{2.58} & \textbf{79.0\%} \\
\midrule
    BART-large &2.05 & 49.5\% & 2.12 & 56.5\%\\
     \plog (BART-large) & \textbf{2.39} & \textbf{67.5\%} & \textbf{2.50} & \textbf{74.5\%}\\
   \bottomrule
    \end{tabular}
    \caption{The human evaluation results of different models. AVG is the average score while ACC means the accuracy of logical fidelity. The average inter-annotator agreement is 0.82 when measured by Fleiss' Kappa. }
    \vspace{-2ex}
    \label{tab:human}
\end{table}

\subsection{Experimental Settings}

\paragraph{Evaluation Metrics} Following prior works~\cite{chen2020logical, chen2021confounded} on \logicnlg, we evaluate our models on both surface-level matching metrics and logical fidelity scores. Surface-level metrics include BLEU-1/2/3, which are based on n-gram matching between the model generations and gold references. In terms of fidelity scores, prior works adopt SP-Acc and NLI-Acc. For SP-Acc, a sentence is first parsed into a logical program and evaluated as the execution accuracy of the program. NLI-Acc is based on TableBERT, a table-entailment model pretrained on the TabFact dataset~\cite{chen2019tabfact}. The model can predict whether a table supports a sentence. 

However, these two fidelity metrics are not enough to verify the fidelity: we empirically find that the parsing algorithm for SP-Acc often generates irrelevant logical programs for the sentences, which renders the evaluation inaccurate. In addition, the TableBERT model used for NLI-Acc only achieves 65.1\% accuracy on the TabFact dataset, and we find it overly positive about the predictions. To this end, we add two state-of-the-art table-entailment models for evaluation: \tapex-large~\cite{liu2021tapex} and \tapas-large~\cite{tapas2020}, which achieve 84.2\% and 81.0\% test accuracy on TabFact, respectively. We name the two metrics as \tapex-Acc and \tapas-Acc, respectively. We still evaluate SP-Acc and NLI-Acc to compare our method with previous studies. For \contlog, we adopt the evaluation metrics of \logictext: BLEU-4 and ROUGE-1/2/4/L to evaluate surface-level matching, and use \tapex-Acc and \tapas-Acc to evaluate the fidelity. 

\paragraph{Models for Comparison}
For \logicnlg, we compare our method with the following models: \textbf{GPT-TabGen (sm)} and \textbf{GPT-Coarse-to-Fine (sm)}~\cite{chen2020logical} are two baselines based on pretrained GPT-2; \textbf{DCVED+GPT-TabGen}~\cite{chen2021confounded} is a de-confounded variational model with GPT-TabGen (sm) as the backbone. We also include pretrained \textbf{BART-large}, \textbf{T5-base} and \textbf{T5-large} as the baselines models for both \logicnlg and \contlog, for which we adopt our data pre-processing method introduced in Section~\ref{sec:model}. Our models are named \textbf{\plog (BART-large)}, \textbf{\plog (T5-base)} and \textbf{\plog (T5-large)} when using different backbones. We adopt the same input serialization strategy with numerical pre-computation for BART, T5, and \plog models.

\paragraph{Training Details}
 
We conduct our main experiments based on Transformers~\cite{transformers} and PyTorch~\cite{pytorch}. During training, the parameters of embedding layers of models are frozen. During inference, we adopt beam search with beam size 4 for all the experiments. We set the maximum length as 500 and 200 for source and target sequences, respectively. Each experiment was run only once because of the time cost. On \logicnlg, model selection is based on the BLEU-3 score on the validation set, and on \contlog, it is based on validation BLEU-4 scores. The selection of pretraining checkpoints is based on the Execution Accuracy of generated logical forms on the validation set of pretraining tasks. We provide detailed hyperparameters in Appendix~\ref{sec:setting}.

\subsection{Automatic Evaluation}

\paragraph{\logicnlg} Table~\ref{tab:lnlg-res} presents the results on \logicnlg. We can observe that the BART and T5 models with our preprocessing strategies outperform all the previous models based on GPT-2 in terms of both surface-level metrics and logical fidelity scores. We also observe that the \plog models mostly outperform their base models on BLEU scores while they can significantly improve the logical fidelity scores on all the metrics. For example, \plog (T5-large) improves the \tapex-Acc and \tapas-Acc over T5-large by an average of 10\% accuracy. However, \plog (T5-base) achieves lower results on BLEU scores, possibly because of the uncontrollable task setting of \logicnlg.
\logicnlg does not provide highlighted cells, so the potential space for content selection is usually very large. This makes models very likely to generate faithful sentences that describe different facts/contents from the gold references, causing low BLEU scores. Moreover, BLEU is based on local N-Gram matching which cannot capture the global faithfulness of generated sentences. Therefore, such surface-level metrics may not correlate well with fidelity metrics.

\paragraph{\contlog} The results on \contlog are shown in Table~\ref{tab:clog-res}. As observed, \plog models outperform their base counterparts consistently on both surface-level and logical-level metrics. This suggests that adding highlighted cells to narrow down the scope of content selection is beneficial to more reliable evaluation. In addition, the consistent improvements with different backbone models demonstrates the general effectiveness of our approach.

\subsection{Human Evaluation}
To further investigate whether the models can generate faithful sentences, we perform a human evaluation on the outputs of BART, T5, and \plog models. Specifically, we randomly sample 200 examples from the test set of each dataset. We hire three human annotators to rate each sentence a score in the discrete range between 0 and 3, according to the criteria adopted in \cite{chen2020logical}. Non-sense (0): the sentence does not make sense, and people cannot understand its meaning. Wrong (1): the sentence is overall fluent, but the logic it describes is false. Partially correct (2): the sentence describes multiple facts. At least one of them is correct, but it still contains factual errors. Correct (3): the sentence is of high quality in both fluency and logical correctness. The model names are hidden to the annotators, and we collect their individual results to summarize two scores for each model: (1) the average of their scores on each sampled set; (2) the fidelity accuracy, i.e., the proportion of sentences scored as correct\footnote{We take a vote on the three evaluators' scores, i.e., a sentence is judged as correct if at least two of them give a score of 3.}. The evaluation is only based on the context table without considering gold references, because the generated sentences may not describe the same fact as the references do but still present high quality in terms of fidelity and fluency.

As shown in Table~\ref{tab:human},
\plog (T5-base) outperforms T5-base by a large margin on \contlog while it does not achieve superior results on \logicnlg, which is inconsistent with automatic scores. However, \plog (T5-large) and \plog (BART-large) achieve significant improvements over base models on both datasets, showing an improvement consistent with the automatic metrics.

\begin{table}[t]
    \centering
    \small
    \begin{tabular}{lcccc}
    \toprule
    \multirow{2}{*}{Model}     & \multicolumn{2}{c}{\logicnlg} & \multicolumn{2}{c}{\contlog} \\
    \cmidrule(lr){2-3} \cmidrule(lr){4-5}
    & Val & Test & Val & Test\\
    \midrule
    \plog (BART-large)   &49.47  & 49.85 & 59.67 & 61.73\\
   \plog (T5-base)     & 90.93 & 88.86 & 91.87  & 92.20 \\
   \plog (T5-large)     & 93.77& 92.23 &  93.33 & 93.13\\
 
   \bottomrule
    \end{tabular}
    \caption{Experimental results of different \plog models on the validation and test sets of table-to-logic generation. The scores are reported as Execution Accuracy.}
    \label{tab:pt-res}
    \vspace{-2ex}
\end{table}

\begin{figure}[t]
    \centering
    \includegraphics[width=\linewidth]{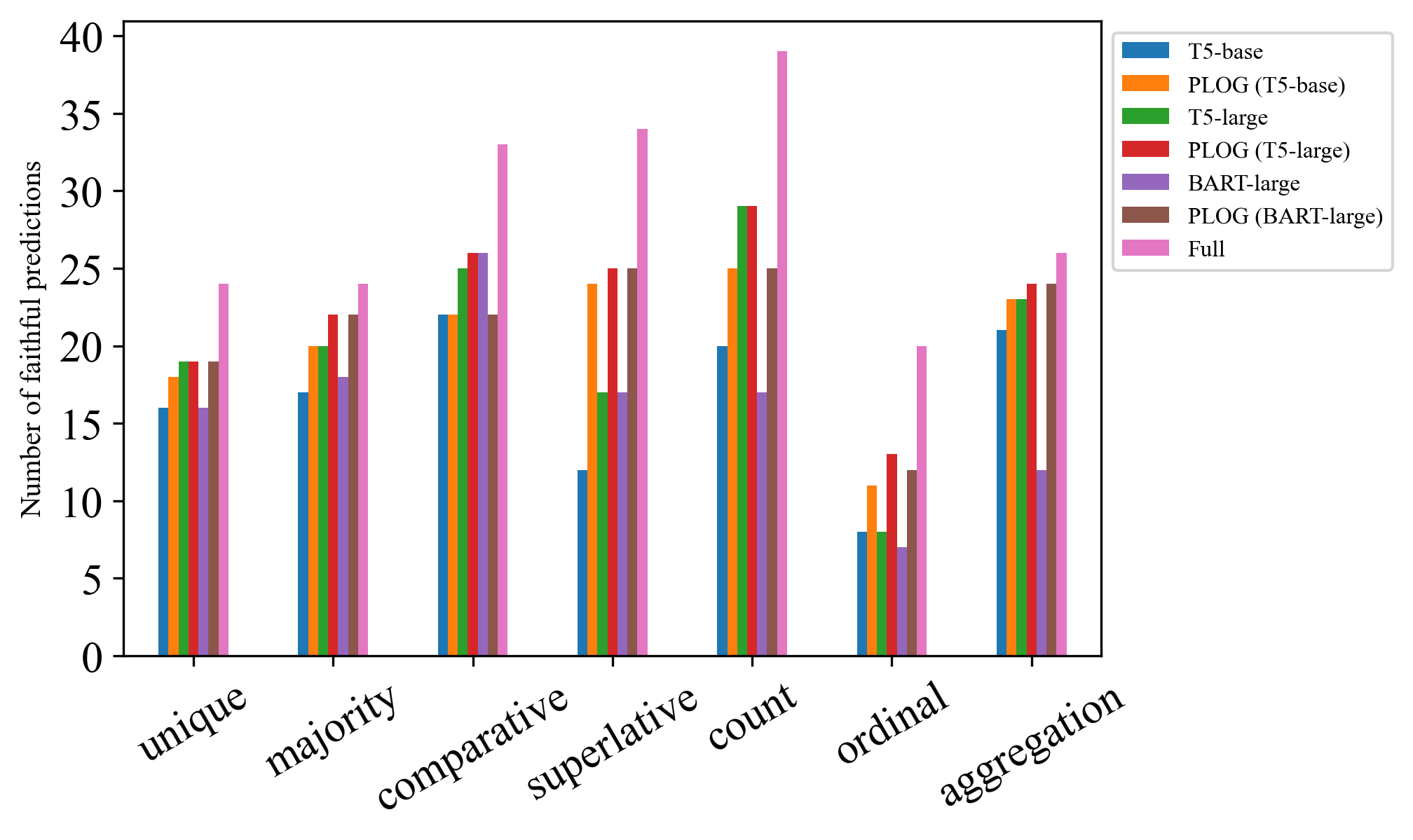}
    \caption{The human evaluation results of different models on the different logic types of \contlog. The y-axis indicates the number of samples scored as correct. \textit{Full} indicates the number of samples of each logic type in the 200 human evaluation samples. }
    \vspace{-1ex}
    \label{fig:type}
\end{figure}

\subsection{Table-to-Logic Results} 
We report the Execution Accuracy of our pretrained models on the table-to-logic pretraining task in Table~\ref{tab:pt-res}. As shown, \plog (T5-base) and \plog (T5-large) present over 90\% accuracy in generating correct logical forms, demonstrating that table-to-logic pretraining indeed improves the model's ability to derive accurate logical facts. However, \plog (BART-large) achieves much lower accuracy. We analyzed the error cases of BART-large and found that over 90\% of the errors are caused by logical form parsing errors, i.e., the generated logic string cannot be successfully parsed into a structurally correct logical form tree because of misspelled function names and mismatched brackets. It seems BART-large performs much worse than T5-base and T5-large at learning the structure of logic strings. We suppose that incorporating grammar-guided decoding methods~\cite{wang2018robust} may alleviate this problem, which we leave to future work. Surprisingly, this does not affect the performance of \plog (BART-large) on downstream tasks, showing that the model still acquired beneficial knowledge through the pretraining.

\subsection{Analysis on Different Logic Types}  
In \contlog, each target sentence belongs to a pre-defined logic type inherited from \logictext, allowing us to analyze the performance of models on different logical reasoning types. In Figure~\ref{fig:type}, we can observe that our \plog models generally improves the performance of their base models on most logic types, especially on \texttt{superlative} and \texttt{ordinal}. However, we still observe a considerable amount of incorrect generations of all the models, suggesting the potential room for improvement in the future.

%% file: Conclusion.tex
\section{Conclusion}
We proposed a table-to-logic pretraining task to enhance the fidelity of logical table-to-text generation. In addition, we constructed a controlled logical table-to-text generation task by re-purposing an existing dataset. To realize pretraining on large-scale corpora, we proposed an execution-guided sampling scheme to extract accurate logical forms from tables automatically. With table-to-logic pretraining, our table-to-text model could significantly improve logical fidelity. Our work shows a novel way to utilize formal language to promote table-to-text generation, and may be extended to other related areas such as table representation learning.

%% file: Appendix.tex
\appendix

\section{Experimental Setting Details}
\label{sec:setting}
 The following are the hyperparameters for different model configurations. During finetuning, each pair of base model and the corresponding \plog model share the same hyperparameters for a fair comparison, while these hyperparameters are tuned only with the base model.

\noindent \textbf{T5-base} and  \textbf{\plog (T5-base)} :
Hyperparamters are the same for both datasets.
\begin{itemize}
    \item Optimizer: AdamW~\cite{loshchilov2017decoupled}.
    \item Learning rate: $2\times 10^{-4}$ for pretraining and $1\times 10^{-5}$ for finetuning.
    \item Batch size: 5 for both pretraining and finetuning.
\end{itemize}
\noindent \textbf{T5-large} and  \textbf{\plog (T5-large)} :
Hyperparamters are the same for both datasets.
\begin{itemize}
    \item Optimizer: AdaFactor~\cite{shazeer2018adafactor}.
    \item Learning rate: $2\times 10^{-4}$ for both pretraining and finetuning.
    \item Batch size: 10 ($2\times 5 $ gradient accumulation steps) for both pretraining and finetuning. 
\end{itemize}
\noindent \textbf{BART-large} and \textbf{\plog (BART-large)}:
\begin{itemize}
    \item Optimizer: AdaFactor for both datasets.
    \item Learning rate: $5\times 10^{-4}$ for pretraining on \logicnlg and $2\times 10^{-4}$ on \contlog; $2\times 10^{-4}$ for fine-tuning on both datasets.
    \item Batch size: 256 ($4 \times$ 64) for pretraining on \logicnlg and 32 ($4 \times 8$) on \contlog; 32 ($4 \times 8$) for fine-tuning on both datasets.
\end{itemize}

The following is the additional information of each pretrained model. 
\begin{itemize}
    \item T5-base: ~220M parameters with 12-layer, 768-hidden-state, 3072 feed-forward hidden-state, 12-heads.
    \item T5-large: ~770M parameters with 24-layer, 1024-hidden-state, 4096 feed-forward hidden-state, 16-heads.
    \item BART-large: ~406M parameters with 24-layer, 1024-hidden-state, 16-heads, 

\end{itemize}

\begin{figure}[t]
\centering

\includegraphics[width=\linewidth]{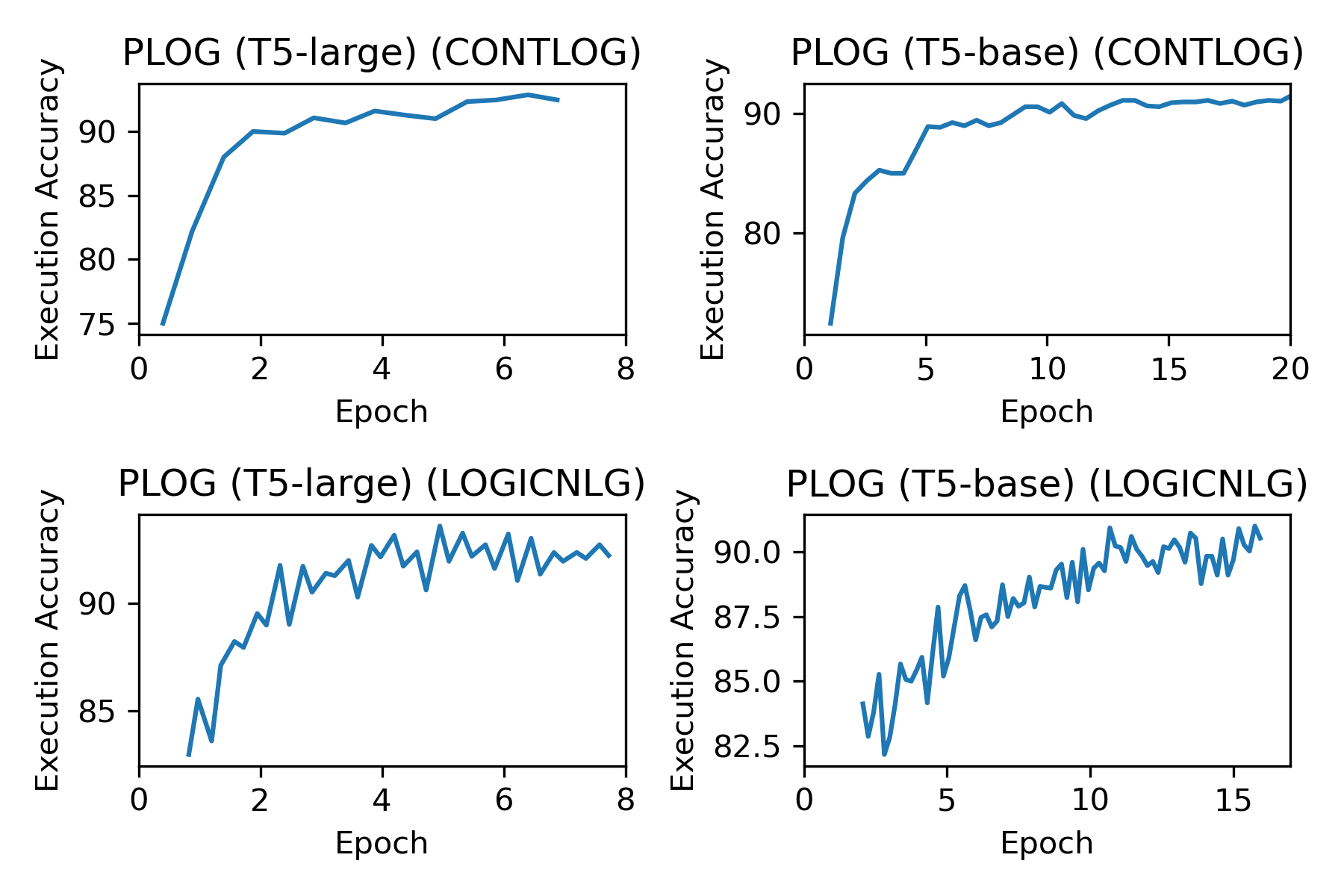}

\caption{Validation results of table-to-logic pretraining with T5-base and T5-large as the backbones. The results of \logicnlg pretraining and \contlog pretraining are shown at different intervals for better illustration. The results within the first 160k steps are not computed. }
\label{fig:curve}
\end{figure}

\paragraph{Pretraining Details}
We pretrain our models on the collected table-to-logic data and evaluate their Execution Accuracy on the validation set (pretraining corpora) at an interval of a certain number of steps. We take the best pretraining checkpoints to finetune them on downstream tasks. 
Figure~\ref{fig:curve} presents the validation results of pretraining during the training process. We can observe that the models achieve higher accuracy when trained for more epochs. The pretraining is very time-consuming because of the large-scale pretraining data and models. For example, it takes approximately 17 hours to train \plog (T5-base) for one epoch on the \contlog pretraining data, while it takes 5 days to train one epoch of \plog (T5-large). Each experiment was done on a single NVIDIA V100 GPU. We suppose the time cost can be reduced by using more GPU resources.

\begin{table*}[ht]
\resizebox{\textwidth}{!}{%
\begin{tabular}{l|l|l|l}
\toprule
Name & Arguments & Output & Description \\
\midrule
count & view & number & returns the number of rows in the view \\
\midrule
only & view & bool & returns whether there is exactly one row in the view \\
\midrule
hop & row, header string & object & returns the value under the header column of the row \\
\midrule
and & bool, bool & bool & returns the boolean operation result of two arguments \\
\midrule
max/min/avg/sum & view, header string & number & returns the max/min/average/sum of the values under the header column \\
nth\_max/nth\_min & view, header string & number & returns the n-th max/n-th min of the values under the header column \\
\midrule
argmax/argmin & view, header string & row & returns the row with the max/min value in header column \\
nth\_argmax/nth\_argmin & view, header string & row & returns the row with the n-th max/min value in header column \\
\midrule
eq/not\_eq & object, object & bool & returns if the two arguments are equal \\
round\_eq & object, object & bool & returns if the two arguments are roughly equal under certain tolerance \\
greater/less & object, object & bool & returns if argument 1 is greater/less than argument 2 \\
\midrule
diff & object, object & object & returns the difference between two arguments \\
\midrule
filter\_eq/not\_eq & view, header string, object & view & returns the subview whose values under the header column is equal/not equal to argument 3 \\
filter\_greater/less & view, header string, object & view & returns the subview whose values under the header column is greater/less than argument 3 \\
filter\_greater\_eq /less\_eq & view, header string, object & view & returns the subview whose values under the header column is greater/less or equal than argument 3 \\
filter\_all & view, header string & view & returns the view itself for the case of describing the whole table \\
\midrule
all\_eq/not\_eq & view, header string, object & bool & returns whether all the values under the header column are equal/not equal to argument 3 \\
all\_greater/less & view, header string, object & bool & returns whether all the values under the header column are greater/less than argument 3 \\
all\_greater\_eq/less\_eq & view, header string, object & bool & returns whether all the values under the header column are greater/less or equal to argument 3 \\
\midrule
most\_eq/not\_eq & view, header string, object & bool & returns whether most of the values under the header column are equal/not equal to argument 3 \\
most\_greater/less & view, header string, object & bool & returns whether most of the values under the header column are greater/less than argument 3 \\
most\_greater\_eq/less\_eq & view, header string, object & bool & returns whether most of the values under the header column are greater/less or equal to argument 3 \\
\bottomrule
\end{tabular}
}
\caption{Function definitions of the logic schema borrowed from \cite{chen2020logic2text}.}
\label{table:app_func}
\end{table*}


\section{Logical Form Schema}
\label{sec:schema}

\logictext~\cite{chen2020logic2text} defines 7 logic types, including \texttt{count, unique, comparative, superlative, ordinal, aggregation} and \texttt{majority}. For the definitions and examples of these logic types, please refer to the Appendix of \cite{chen2020logic2text}. In this section, we provide a complete list of the logical functions in Table~\ref{table:app_func}, which we use to define our logical form schema.

\section{Details of Pretraining Data Collection}
\label{sec:collect}
Here, we provide more details of the pretraining data collection procedure, including examples of abstract templates and the complete rules for logical form sampling. Table~\ref{tab:funcset} lists the function-type placeholders.

\begin{table}[t]
    \centering
    \resizebox{\linewidth}{!}{
    \begin{tabular}{cl}
         \toprule
    Category & Function   \\ \midrule
    FILTER & filter\_eq, filter\_not\_eq, filter\_greater, ... \\
    SUPERLATIVE &  max, min\\
    ORDINAL &  nth\_max, nth\_min \\
   SUPERARG & argmax, argmin \\
   ORDARG &  nth\_argmax, nth\_argmin \\
   COMPARE & greater, less, eq, not\_eq \\
   MAJORITY & all\_eq, all\_not\_eq, most\_eq, all\_greater, ... \\
   AGGREGATE & avg, sum \\
   \bottomrule
    
    \end{tabular}}
    \caption{The categorized functions for template abstraction. Functions in the same category have the same argument definitions. }
    \label{tab:funcset}
\end{table} 


\paragraph{Templates}
We provide in Table~\ref{tab:temp} some examples of our logical form templates. These examples are all based on the example table in Figure~\ref{fig:main}.

\paragraph{Instantiation}
Here we provide the main rules we design for instantiating a logical form template by sampling from a table.
\begin{enumerate}
    \item For placeholder type \texttt{Column}, we randomly sample a column header from the current input (sub-) table.
    \item For placeholder type \texttt{Object}, the instantiation depends on the parent function node of this placeholder. If the function node is \texttt{only} or belongs to the category \texttt{FILTER} or \texttt{MAJORITY}, the placeholder is instantiated as a sampled value from a certain column of the current input (sub-) table. Otherwise, if the function node is \texttt{eq}, this placeholder is instantiated as the execution result of its brother node. This is to guarantee the correctness of equality judgements.
    \item The instantiation of a function-type placeholder depends on its function category, as listed in Table~\ref{tab:funcset}. If the placeholder belongs to the function category \texttt{COMPARE} or \texttt{MAJORITY}, we choose the specific function name based on the real relationships among its arguments. For example, the arguments of \texttt{COMPARE} functions are two objects whose relationship (equal, greater, less, etc.) can be pre-computed. Hence we can determine the actual function based on this relationship. If the placeholder belongs to another category, the function can be uniformly sampled from the function set.

\end{enumerate}

\begin{table*}[t]
  
    \centering{
    \resizebox{\linewidth}{!}{
    \begin{tabular}{lll}
        \toprule
        Logic Type & \multicolumn{2}{c}{Examples} \\
        \midrule
      \multirow{3}{*}{Count}& Template  & 
      
      eq \{ count \{ [FILTER] \{ all\_rows ; [Column 1] ; [Object 1] \} \} ; [Object 2] \}  \\ 
  & Instance & eq \{ count \{ filter\_eq \{ all\_rows ; power ; 20kw \} \} ; 3 \} \\
  & Explanation & In dwbl, there are 3 brandings with power 20kw.\\
  \midrule
  \multirow{3}{*}{Comparative}& Template  & 
  [COMPARE] \{ hop \{ [FILTER] \{ all\_rows ; [Column 1] ; [Object 1] \} ; [Column 2] \} ; hop \{ [FILTER] \{ all\_rows ; [Column 1] ; [Object 2] \} ; [Column 2] \} \}    
       \\ 
  & Instance & greater \{ hop \{ filter\_eq \{ all\_rows ; callsign ; dwbl \} ; power \} ; hop \{ filter\_eq \{ all\_rows ; callsign ; dyku \} ; power \} \}  \\
  & Explanation & The callsign dwbl has a greater power than dyku. \\
   \midrule
  \multirow{3}{*}{Unique}& Template  & 
only \{ [FILTER] \{ all\_rows ; [Column 1] ; [Object 1] \} \}      
       \\ 
  & Instance & only \{ filter\_eq \{ all\_rows ; location ; iloilo city \} \}  \\
  & Explanation & Only one brand is located in iloilo city.\\
   \midrule
  \multirow{3}{*}{Superlative}& Template  & 
    eq \{ hop \{ [SUPERARG] \{ all\_rows ; [Column 1] \} ; [Column 2] \} ; [Object 1] \} 
       \\ 
  & Instance & eq \{ hop \{ argmin \{ all\_rows ; frequency \} ; callsign \} ; dyku \}   \\
  & Explanation & The callsign dyku has the lowest frequency. \\
   \midrule
  \multirow{3}{*}{Ordinal}& Template  & 
     eq \{ hop \{ [ORDARG] \{ all\_rows ; [Column 1] ; [Object 1] \} ; [Column 2] \} ; [Object 2] \}  
       \\ 
  & Instance &  eq \{ hop \{ nth\_argmax \{ all\_rows ; frequency ; 2 \} ; branding \} ; mellow 957 \}\\
  & Explanation & Mellow 957 is the brand that has the second highest frequency.\\
   \midrule
  \multirow{3}{*}{Majority}& Template  & 
      [MAJORITY] \{ all\_rows ; [Column 1] ; [Object 1] \}
       \\ 
  & Instance & most\_less \{ all\_rows ; frequency ; 1242khz \} \\
  & Explanation & Most of the brands have a frequency lower than 1242khz.\\
   \midrule
  \multirow{3}{*}{Aggregation}& Template  & 
  round\_eq \{ [AGGREGATE] \{ all\_rows ; [Column 1] \} ; [Object 1] \}
       \\ 
  & Instance & round\_eq \{ avg \{ all\_rows ; power \} ; 16kw \} \\
  & Explanation & The average power of all the brands is 16kw. \\

       \bottomrule
    \end{tabular}}}
    \caption{Examples of logical form sampling. For each logic type, we show an example of the abstract template, an instance sampled from the table in Figure~\ref{fig:main}, and a textual explanation of the instance.}
    \label{tab:temp}
\end{table*}

 \section{Case Study}
 \label{sec:case}
\begin{figure*}[t]
\centering
\subfloat[An example of \contlog. The yellow-colored cells are highlighted cells.]{%
\includegraphics[width=\linewidth]{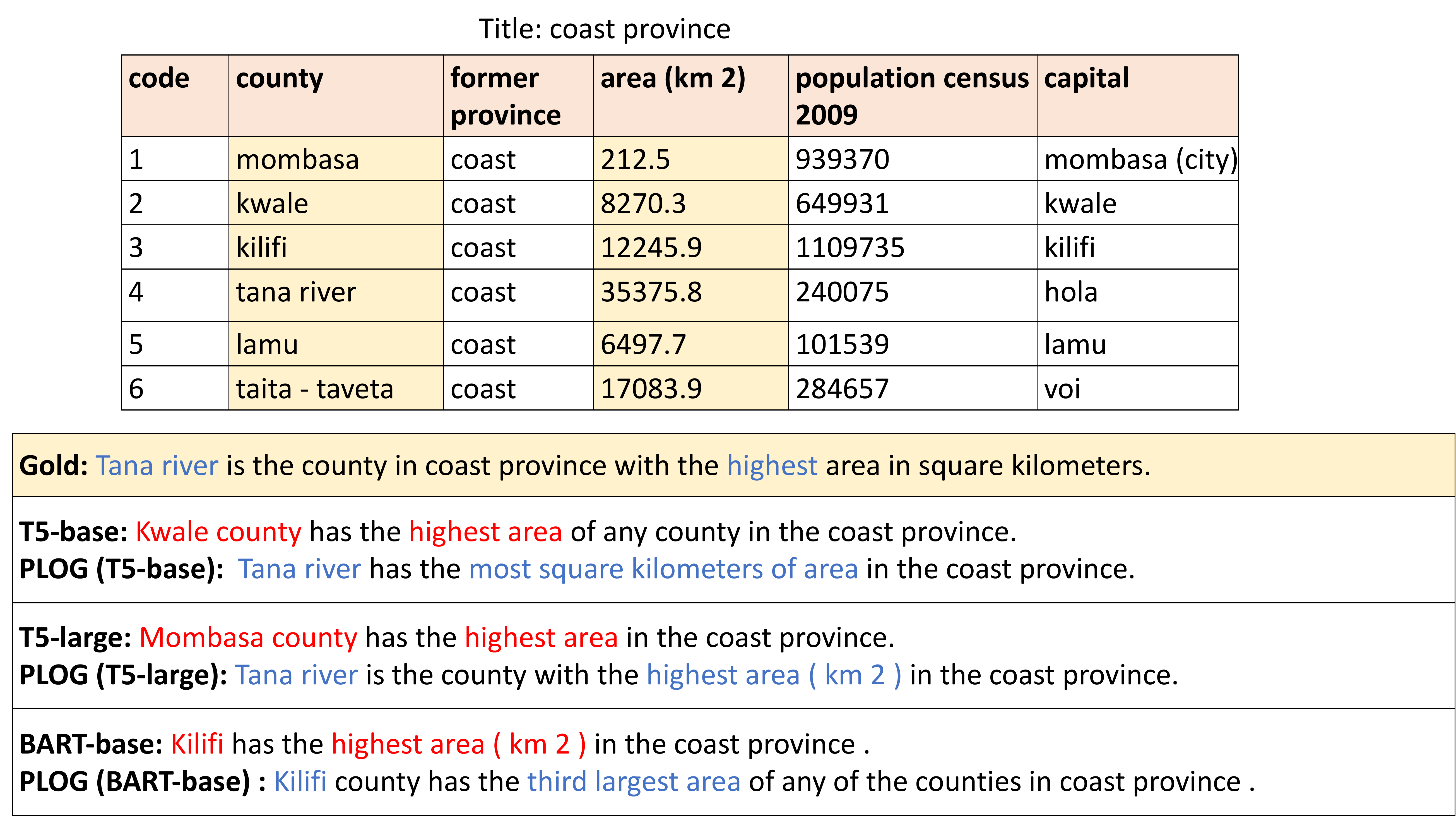}
\label{fig:case1}}

\subfloat[An example of \logicnlg. Some irrelevant columns are removed for illustration.]{%
\includegraphics[width=\linewidth]{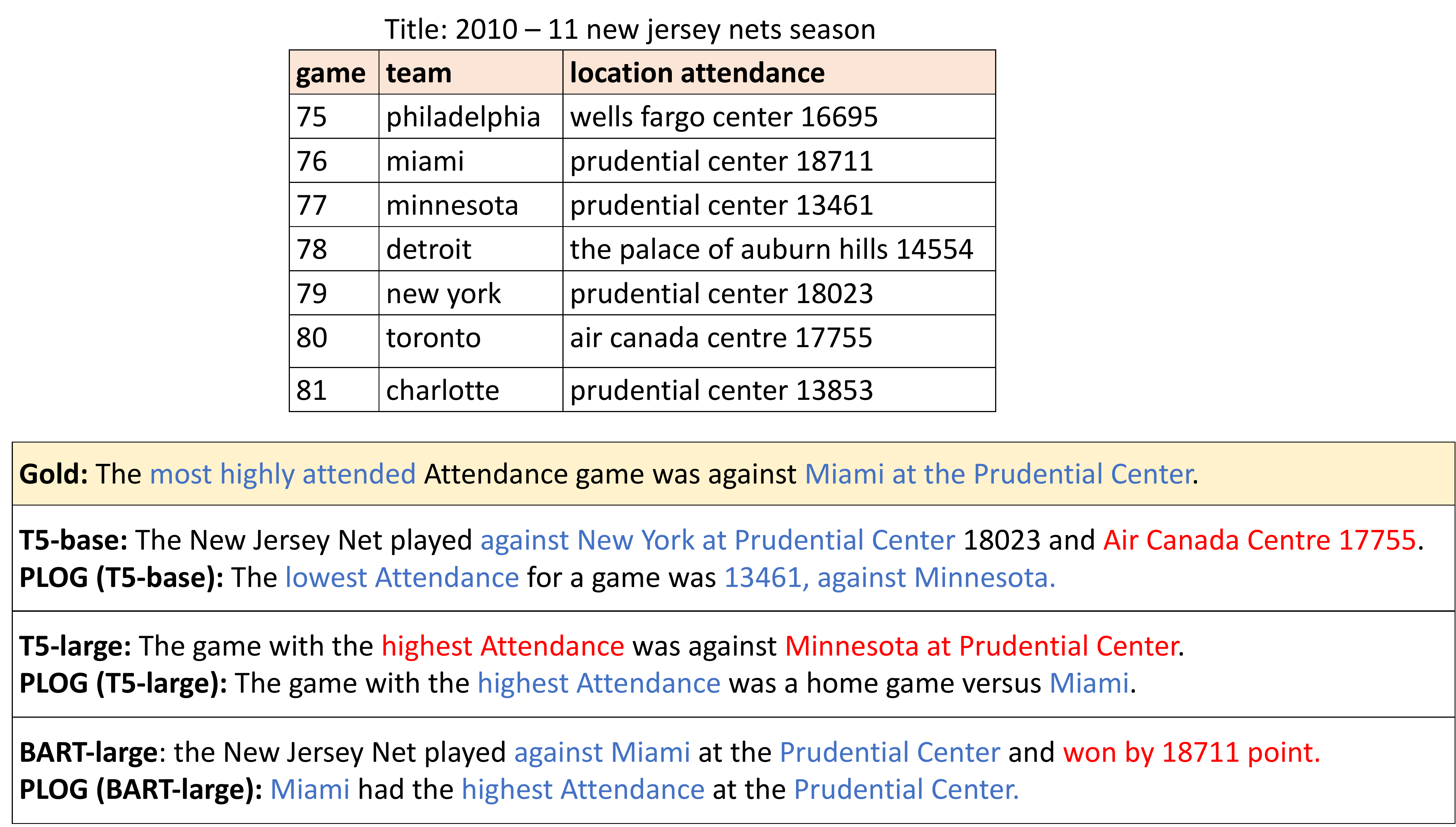}
\label{fig:case2}}

\caption{Qualitative examples of two datasets. The red color indicates incorrect facts while the blue color indicates correct facts.}
\label{fig:case}
\end{figure*}

 We further conduct a case study by showing some qualitative examples of model generations. As presented in Figure~\ref{fig:case}, \plog models can generate logically correct sentences with complex reasoning while the base models often fail to describe correct facts for the table.

